\documentclass[lettersize,journal]{IEEEtran}
\usepackage{amsmath,amsfonts}
\usepackage{algorithmic}
\usepackage{algorithm}
\usepackage{array}
\usepackage[caption=false,font=normalsize,labelfont=sf,textfont=sf]{subfig}
\usepackage{textcomp}
\usepackage{stfloats}
\usepackage{url}
\usepackage{verbatim}
\usepackage{graphicx}
\usepackage{xcolor}
\usepackage[
        numbers,square,
        sort
        ]{natbib}

\definecolor{mycolor}{HTML}{000000}
\usepackage[
        breaklinks,
        colorlinks,
        citecolor=blue,
        linkcolor=blue,
        ]{hyperref}
\begin{document}

\title{Collaborative Perception for Connected and Autonomous Driving: Challenges, Possible Solutions and Opportunities}

\author{Senkang Hu,~\IEEEmembership{Graduate Student Member,~IEEE,} Zhengru Fang,~\IEEEmembership{Graduate Student Member,~IEEE,}\\ Yiqin Deng,~\IEEEmembership{Member,~IEEE,} Xianhao Chen,~\IEEEmembership{Member,~IEEE,}  and Yuguang Fang,~\IEEEmembership{Fellow,~IEEE} 

\thanks{The research work described in this paper was conducted in the JC STEM Lab of Smart City funded by The Hong Kong Jockey Club Charities Trust under Contract 2023-0108. The work was supported in part by the Hong Kong SAR Government under the Global STEM Professorship and Research Talent Hub. The work of Senkang Hu was supported in part by the Hong Kong Innovation and Technology Commission under InnoHK Project CIMDA. The work of Yiqin Deng was supported in part by the National Natural Science Foundation of China under Grant No. 62301300. The work of Xianhao Chen was supported in part by the Research Grants Council of Hong Kong under Grant 27213824. \textit{(Corresponding Author: Yiqin Deng)}}

\thanks{Senkang Hu, Zhengru Fang, Yiqin Deng, and Yuguang Fang are with the Hong Kong JC STEM Lab of Smart City and the Department of Computer
Science, City University of Hong Kong, Kowloon Tong, Hong Kong. (Email: {\{senkang.forest, zhefang4-c\}@my.cityu.edu.hk, \{yiqideng, my.Fang\}@cityu.edu.hk}) 
}
\thanks{Xianhao Chen is with the Department of Electrical and Electronic Engineering, The University of Hong Kong, Pok Fu Lam, Hong Kong. (Email:{ xchen@eee.hku.hk})}
}



\maketitle
\begin{abstract}
Autonomous driving has attracted significant attention from both academia and industries, which is expected to offer a safer and more efficient driving system. 
However, current autonomous driving systems are mostly based on a single-agent perception, which has significant limitations, causing serious safety concerns. Collaborative perception with connected and autonomous vehicles \textcolor{mycolor}{(CAV)} shows a promising solution to overcoming these limitations. In this article, we first identify the challenges of collaborative perception, such as data sharing asynchrony, large data volume, and pose errors. Then, we discuss the possible solutions to address these challenges with various technologies, where the research opportunities are also elaborated. Furthermore, we propose a scheme to deal with communication efficiency and latency problems, which is a channel-aware collaborative perception framework to dynamically adjust the communication graph and minimize latency, thereby improving perception performance while increasing communication efficiency.
Finally, we conduct experiments to demonstrate the effectiveness of our proposed scheme.


\end{abstract}


\section{Introduction}






\IEEEPARstart{W}{ith} the development of artificial intelligence and computationally efficient hardware, autonomous driving has attracted significant attention from both academia and industries, which is expected to bring a safer and more efficient driving system \cite{hanCollaborativePerceptionAutonomous2023a,fang2024pacp}. However, current autonomous driving system is mostly based on a single vehicle, i.e., a single-agent perception system, which has significant limitations that may threaten the driving safety. For example, occlusion is an inevitable problem that a single-agent perception system cannot overcome, which may cause serious accidents. In addition, the sensing range of a single-agent perception system is limited by the deployed sensors (\textit{e.g.}, the maximum detection distance of LiDAR and the field of view of camera), which may miss some important information. As a result, we need to develop a more robust perception system to address these limitations.
Recently, vehicle-to-vehicle (V2V) and vehicle-to-everything (V2X) communications have emerged as viable technologies for autonomous driving, which opens up a new field of research, namely, collaborative perception with connected and autonomous vehicles. Collaborative perception with CAVs shows a promising solution to tackling the limitations on a single-agent perception system, which can improve the reliability and safety of autonomous driving. 
In addition, it presents a wealth of opportunities, including an improved user experience, heightened road safety, better air quality, and a range of innovative transportation solutions. However, to effectively implement this technology, several critical design issues need to be addressed. These include concerns about data privacy and ethics, the development of smart city infrastructure, standardization and regulatory policies from governments and regulatory bodies, building trust among humans and the system, and finding the right balance between over-reliance and communication overhead. Therefore, while the benefits of collaborative perception are abundant, a comprehensive and multifaceted approach is essential for its successful integration into everyday life.

In order to investigate collaborative perception, some public large-scale datasets for collaborative perception have been released. For example, Xu \textit{et al.} proposed OPV2V \cite{xuOPV2VOpenBenchmark2022}, which is an open dataset for collaborative perception, supporting LiDAR-based 3D object detection and camera-based bird's eye view (BEV) segmentation. 
Moreover, there are many studies dedicated to designing collaboration modules that aim to balance spectrum bandwidth and perception accuracy. These focus on addressing issues such as with whom to communicate, when to communicate, how to share information, and how to fuse and aggregate features, among others.
\textcolor{mycolor}{However, these works typically assume an ideal scenario that does not consider practical autonomous driving environments. For example, Where2comm \cite{huWhere2commCommunicationefficientCollaborative2024} assumes that information sharing is synchronous, but in practice, it may be asynchronous. In addition, most of these works assume that all the vehicles are benign \cite{huWhere2commCommunicationefficientCollaborative2024,wangV2VNetVehicletoVehicleCommunication2020a,xuV2XViTVehicletoEverythingCooperative2022a}, but in practice, some vehicles may be compromised, then become malicious and attack the collaborative perception system. These practical issues pose serious challenges for realizing reliable collaborative perception.}
In this article, we investigate the challenges in collaborative perception and elaborate possible solutions based on various technologies, and we also point out some future research opportunities. Finally, we leverage an exemplar scheme to address some of the challenges in collaborative perception and conduct experiments to demonstrate the effectiveness of our proposed scheme.

\section{Background}
\label{sec:background}
\subsection{Demands for Autonomous Driving}

A major factor driving the development of autonomous vehicles is the belief that they will reduce the number of accidents, injuries, and fatalities compared to vehicles driven by humans. 
However, self-driving autonomous vehicles may make incorrect decisions because of errors in detecting and recognizing objects. 
This can be compared to the poor choices a human driver might make under the influence of alcohol or fatigue.
Decisions made by autonomous vehicles due to these detection failures could be as detrimental, if not more so, as those made by their human counterparts. 
For example, in a crash involving a Tesla in California, the vehicle made a critical error, resulting in a fatal outcome: its sensors detected the concrete barrier but dismissed this information, mistakenly interpreting the barrier's stationary position on the radar as non-threatening. Additionally, various errors made by autonomous vehicles have also led to accidents, resulting in loss of life and property. 
As a result, the safety of autonomous driving has become a significant public concern, underscoring the need for the development of safer autonomous driving.

\subsection{The Limitations on a  Single-Agent Perception}

Although the single-agent perception has made great progress in many vision-based tasks in recent years, such as 2D/3D object detection, semantic segmentation, BEV segmentation, and tracking, it still has some limitations. The first is occlusion and the second is the limitation of sensing range. 

\begin{figure}[t]
  \centering
  \includegraphics[width=0.7\linewidth]{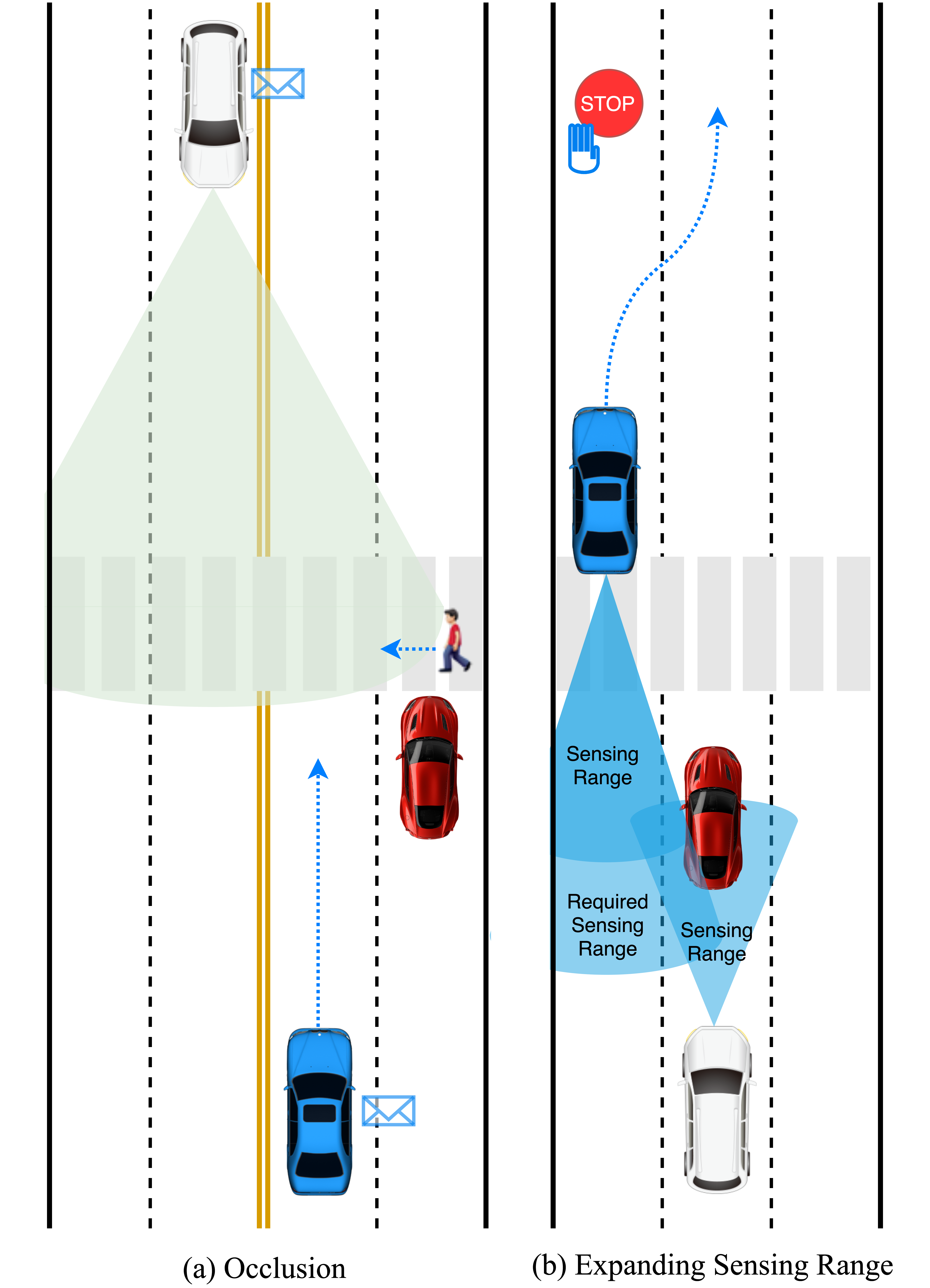}
  \caption{The visualization of occlusion and the limitation of sensing range.}
  \label{fig:example_occlusion}
\end{figure}

\textbf{Occlusion.} Occlusion is a common phenomenon in the real world, where objects are blocked by other objects. In case of occlusion, a single-agent perception system may not detect the occluded objects, potentially leading to safety hazard. For example, as shown in Fig. \ref{fig:example_occlusion}(a), a person crosses the road, but he/she is blocked by the red vehicle from the blue vehicle's viewpoint. This poses a significant risk of an accident if the blue vehicle continues moving forward. This scenario highlights a critical limitation of single-agent perception systems. However, a white vehicle in the opposite lane has a clear view of the person. If the white vehicle can share the information with the blue vehicle, it will enable the blue vehicle to take preventive action and avoid a potential accident.

\textbf{Limitation of Sensing Range.} 
The sensing range of a single-agent perception system is limited by the capabilities of its sensors, such as cameras and LiDAR. For instance, a camera's sensing range is confined to its field of view (FOV), and a LiDAR's range is restricted by its maximum detection distance (\textit{e.g.}, off-the-shelf LiDAR typically has a sensing range of 90 meters). These limitations may result in the omission of some objects, potentially leading to safety hazards. For example, as illustrated in Fig. \ref{fig:example_occlusion}(b), blue vehicle intends to change lanes into a high-speed traffic lane. The time required for the lane change and the acceleration duration determine the necessary perception range behind blue vehicle to execute the maneuver safely. When sensor data is shared, vehicles approaching from behind, like white vehicle, can enhance blue vehicle's perception by extending its sensing range.

\subsection{Development of V2X Communications}



The initial technologies for vehicular communications primarily encompass vehicle-to-vehicle (V2V) and vehicle-to-infrastructure (V2I) modes. V2V communications enable vehicles to exchange information amongst themselves, whereas V2I communications facilitate information access via roadside units (RSUs). With the advent of Internet of Things (IoT) technologies, the concept of vehicle-to-everything (V2X) communications has emerged, which is crucial for enabling collaborative perception in autonomous driving. 
{For example, dedicated short-range communications (DSRC) technology is specifically designed for vehicular communications, which can enable V2V and V2I communications. In addition, 5G NR-V2X is another promising technology to support both cellular and direct communications, including sidelink communications. However, the development of V2X communications faces several challenges, such as high mobility, dynamic network topology, and frequent handovers, which can result in high latency and packet loss.}
In order to tackle these challenges, 6G is set to play a pivotal role in future CAVs. 6G-enabled V2X communication promises to provide services with ultra-low latency, high data transmission rates, and reliable connections, with greater energy and spectrum efficiency than previous generations of wireless communications.

\section{Design Challenges}\label{sec:challenges}

In order to develop a reliable collaborative perception system for autonomous driving, there are a lot of design challenges that need to be addressed, such as large data volume, asynchronous information sharing, collaborative security, and pose errors. In this section, we will discuss these challenges in detail.

\textbf{Large Data Volume.} In autonomous vehicles equipped with different sensors, such as camera, LiDAR, and radar.
the volume of sensing results (\textit{e.g.}, point clouds, image sequences, etc.) is very large. For example, according to KITTI dataset \cite{geigerAreWeReady2012}, each frame produced by 3D laser scanners contains around 100,000 points. The smallest scene captured in this dataset includes 114 frames, resulting in a total of more than 10 million points. In order to realize real-time, efficient collaboration, how to share the useful information in the large volume of data is a grand challenge.

\textbf{Asynchronous Information Sharing.} 
In an LTE-V2X communication system, the average latency can reach as much as 131.30 ms. Additionally, the differing latency from nearby CAVs to ego CAV across various communication channels can cause serious problems with data fusion.
Specifically, in a collaborative perception system, CAVs need to share information with ego CAV, due to the large volume of data and the limited transmission rate, there is a time-lapse from the moment that other CAVs start sending information to the moment that the ego CAV receives the information. This time-lapse results in information sharing at different time instants, causing the ego CAV to fuse information at different instants rather than at the same time. 
Through experimentation, it has been observed that this information sharing latency problem greatly impairs the performance of a collaborative perception system, resulting in performance even inferior to that of a single-agent perception system.


\textbf{Pose Errors.} For effective collaboration, it is essential that multiple CAVs share accurate poses with each other. This necessitates the synchronization of their individual data within a uniform spatial coordinate system, forming the foundation for their cooperative efforts. However, in practical scenarios, the six Degrees of Freedom (DoF) pose estimated by each CAV's localization system is not always precise. This lack of precision causes unintended errors in relative pose estimation, which can significantly diminish the collaborative effort.

\textbf{Collaboration Security.} Collaboration security is another challenge that needs to be addressed. Compared with a single-agent perception system, a collaborative perception system faces higher susceptibility to adversarial attacks. In collaborative perception, each CAV uses an identical model to independently transform local sensor inputs into feature maps. These CAVs, part of the same collaborative network, then exchange these feature maps. Each CAV combines the received feature maps with its own, significantly enhancing performance in subsequent tasks. However, this exchange of feature maps between CAVs also opens up possibilities for adversaries to compromise the entire system. For instance, an attacker could engage in a man-in-the-middle attack, altering the feature maps sent to the target CAV, or a malevolent CAV might directly transmit tampered feature maps to a victim. This is particularly difficult to detect  if the feature maps are generated from adversarial perceptual information such as high-definition maps. Additionally, since humans cannot visually interpret the encoded feature maps, even minor modifications to these maps can go undetected, making such attacks particularly covert. This vulnerability hinders the deployment of advanced collaborative perception tools in safety-critical environments. For instance, in autonomous driving, any failure in perception can lead to severe outcomes, including property damage or, in the worst case, loss of human lives.

\section{Possible Solutions and Opportunities}\label{sec:countermeasures}

In this section, we will discuss the possible solutions to address the aforementioned challenges, and we also investigate the future research opportunities, including multi-modal collaboration, generalization to real world scenarios, and collaboration security issues.
\subsection{Possible Solutions}

\textbf{Communication Efficiency. 
} Due to the large volume of sensing data, it is not feasible for nearby collaborative vehicles to transmit all the sensing data. Therefore, efficient methods are needed for fast and efficient communications between CAVs to enable efficient collaborative perception.

In general, there are three methods to share information through V2V communications: 1) early fusion, 2) intermediate fusion, and 3) late fusion. Early fusion means that the raw sensed data (\textit{e.g.}, raw point clouds, raw camera images) from nearby collaborative vehicles is transmitted to  ego vehicle and are then fused with ego vehicle's own sensed data to perform perception tasks. Intermediate fusion fuses intermediate features extracted from the raw data. 
Late fusion aggregates the final results of perception tasks, such as bounding boxes and BEV maps. In this method, each vehicle first performs perception tasks on its own sensing data, then transmits the final results to the ego vehicle, which combines these results to obtain the final outcome.

Early data fusion is the most straightforward method, but it has the highest communication overhead. The late data fusion has the lowest communication overhead, but it may result in information loss. Recent research \cite{huAdaptiveCommunicationsCollaborative2023} indicates that intermediate fusion may yield the optimal trade-off between perception accuracy and communications. 

There already exist some research along this line. Wang \textit{et al.} \cite{wangV2VNetVehicletoVehicleCommunication2020a} leveraged a fully-connected graph neural network (GNN) to aggregate the intermediate features received from other CAVs. They first calculate the relative poses between other CAVs and ego vehicle, then warp the intermediate features to the ego vehicle's coordinate system and send them to the ego vehicle. Secondly, they use a mask-aware accumulation operation to aggregate the intermediate features while ensuring only overlapping fields-of-view is considered. Hu \textit{et al.} \cite{huWhere2commCommunicationefficientCollaborative2024} proposed a spatial confidence-aware message fusion, which leverages multi-head attention to fuse the corresponding features from multiple CAVs at each individual spatial location. A key aspect of this design is the use of spatial confidence maps. These maps show how confident each CAV is about its data at different locations. By including these feature maps, the system can better learn and focus on the most reliable information from each CAV, enhancing the overall decision-making process.



\textbf{Asynchronous Information Sharing.} 
Due to the differences of sensing instants from different CAVs and the channel variations between collaborative CAVs and the ego vehicle, shared information may arrive at the ego vehicle asynchronously. This is a critical issue in collaborative perception, which may cause the loss of some information.
The main idea to tackle this problem is to compensate the time delay for either sensing or communication. For example, Wang \textit{et al.} \cite{wangV2VNetVehicletoVehicleCommunication2020a} proposed a method to compensate the time delay. Specifically, they constructed a graph in which each node represents the state representation of a CAV. They first initialize the state of each node in this graph. For every node, the intermediate features received from other CAVs, the relative six DoF poses, and the time delay relative to the sensing time of the receiving vehicle are fed into 
a convolutional neural network (CNN). This process yields a time-delay-compensated representation, effectively adjusting for the discrepancies in data acquisition timing among CAVs.
Lei \textit{et al.} \cite{leiLatencyAwareCollaborativePerception2022a} proposed a latency-aware framework, which utilizes past collaboration data to concurrently estimate the present feature and the associated collaboration attention, both of which are unknown because various kinds of the latencies. This framework can dynamically adjust the asynchronous perceptual features from multiple agents to align with the same time stamp, thereby enhancing the robustness and effectiveness of collaborative perception.

\textbf{Pose Errors.} In order to address pose errors, there are two main methods: 1) deep learning based methods, 2) pose correction methods. Xu \textit{et al.} \cite{xuV2XViTVehicletoEverythingCooperative2022a} proposed an attention mechanism to handle pose errors, named multiscale window attention. This approach employs a hierarchical structure of windows, each with a unique attention span. By incorporating windows of varying sizes, the robustness of detecting objects is significantly enhanced, especially in face of localization inaccuracies. Lu \textit{et al.} \cite{luRobustCollaborative3D2023a} proposed a pose correction method before transmitting the information. Specifically, they constructed an agent-object pose graph and optimized it, which could align the relative pose relationships between CAVs and the detected objects in the scene, thereby promoting the pose consistency. In addition, they proposed a multiscale intermediate fusion strategy to aggregate the information at multiple spatial scales, which could further alleviate the impact of pose errors.

\textbf{Collaboration Security.}
Zhao \textit{et al.} \cite{zhaoMaliciousAgentDetection2023} proposed a malicious CAV detection method, which can detect the evasion attacks against object detectors that aim to degrade the performance of a machine learning system at inference time. In particular, this approach utilizes the dynamics of CAV collaboration to effectively identify and eliminate malicious CAVs within an ego CAV's collaboration network. It employs a dual-testing strategy to regulate the false positive rate. The testing involves two separate hypothesis evaluations, focusing on the consistency between the ego CAV and the CAV under inspection, with the baseline assumption that they are ``consistent". They introduced two innovative detection metrics for these tests: 1) a match loss statistic assessing the alignment between the bounding box proposals from both CAVs, and 2) a collaborative reconstruction loss statistic evaluating the congruence of their fused feature maps. A CAV is deemed malicious if it is found ``inconsistent" with the ego CAV based on these tests. This method is distinct from proactive strategies as it not only shields CAVs from potential adversarial threats but also aids in identifying the sources of such attacks, which has significant social implications. Methodologically, this method combines the strengths of both detection metrics while maintaining a controlled false positive rate, and it is designed to be expandable to incorporate additional detection statistics from future research.

\begin{figure*}[h]
  \includegraphics[width=\textwidth]{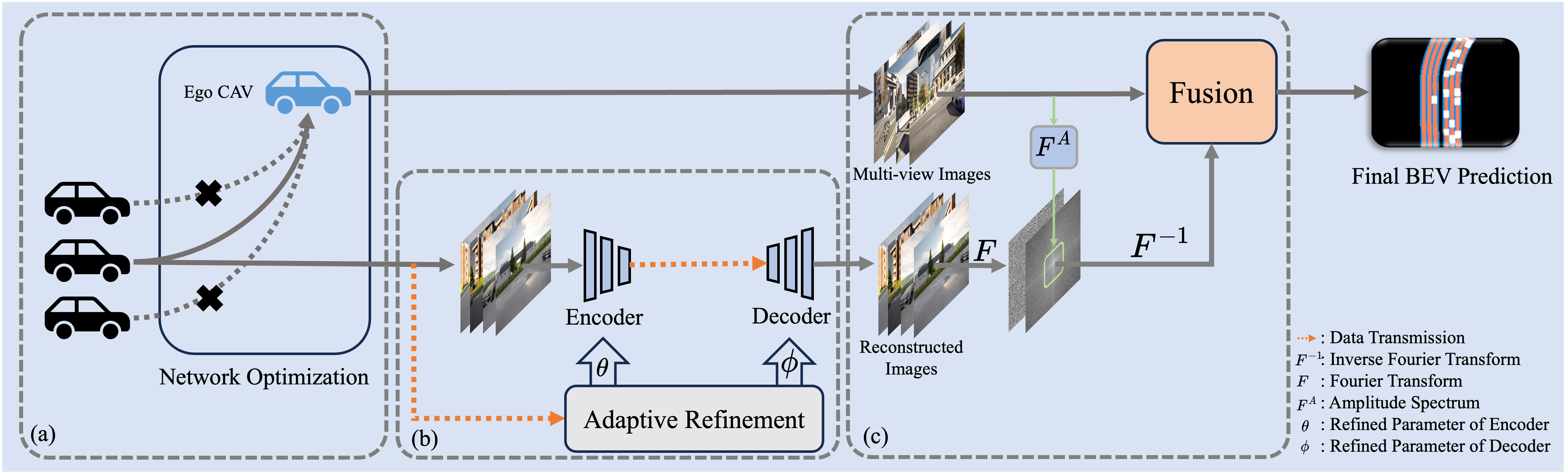}
  \caption{Overall architecture of the proposed collaborative perception framework.}

  \label{fig:overall-architecture}
\end{figure*}

\subsection{Opportunities}

\textbf{Multi-Modal Collaboration.} \textcolor{mycolor}{Most current works consider a homogeneous scenario where all CAVs use identity sensors and perception models. However, in reality, different CAVs may be equipped with different sensors (e.g., LiDAR, camera, radar). For example, if one CAV is equipped with LiDAR and camera while another CAV is just equipped with camera, how can they perform the collaborative perception? In order to address this issue, a more powerful multi-modal collaborative perception system that can handle the heterogeneous CAVs is needed.
This is an open problem and needs to be investigated in the future. }

\textbf{Generalization to Real Scenarios.} \textcolor{mycolor}{Constructing real-world datasets for collaborative perception is challenging. Up to now, there is only one open-source dataset collected from real-world scenarios for vehicle-to-vehicle (V2V) collaborative perception, namely, V2V4Real \cite{xuV2V4RealRealWorldLargeScale}. However, this dataset is limited and only has 2 CAVs in each scene. As a result, many collaborative perception approaches depend on simulated datasets for their development, but the models trained on simulated data may not generalize well to real-world collaborative environments.} Therefore, it is crucial to investigate effective domain adaptation techniques \cite{huAdaptiveCommunicationsCollaborative2023}. Such techniques will facilitate the transition of collaborative perception models from simulated to real environments, and vice versa, thereby enhancing the applicability of V2X collaborative perception in diverse real-world settings.

\textbf{Collaborative Perception with Security Consideration.} \textcolor{mycolor}{As collaborative perception is a relatively new topic, there are few works that consider the security issues in this area. Additionally, since collaborative perception is a distributed system, the attack surface is larger, which significantly weakening cyber defenses and making the system more vulnerable to malicious attacks from untrustworthy participants.} Therefore, addressing these security concerns necessitates a shift in focus in future research towards enhancing trust among collaborating agents, which involves not only recognizing and mitigating malevolent or self-interested behaviors but also integrating these considerations into collaborative perception strategies.

\textcolor{mycolor}{\textbf{Willingness and Incentives.} In collaborative perception,  some vehicles may not be willing to share their data due to privacy concerns or other reasons. Therefore, how to stimulate vehicles to participate in the collaborative perception by designing an effective incentive mechanism is a research opportunity worth exploring.
}

\section{Case Studies} \label{sec:example-scheme}

As discussed above, there are many design challenges in collaborative perception that we need to address carefully. \textcolor{mycolor}{In this section, we propose an exemplar collaborative perception framework. This framework not only minimizes the overall transmission latency but also address the domain misalignment issue between different CAVs. Compared with other baseline designs, our method achieves the best performance.} {In addition, the scheme proposed in this section provides a high-level idea, and the technical implementation and the full evaluation are discussed in \cite{huAdaptiveCommunicationsCollaborative2023,fang2024pacp}.}

\subsection{System Architecture}

The overall architecture of the proposed collaborative perception framework is shown in Fig. \ref{fig:overall-architecture}. The proposed framework consists of three main components: 1) transmission delay minimization module, 2) adaptive data reconstruction module, and 3) domain alignment module. 

In this framework, our initial step involves optimizing the communication network by adjusting it according to the dynamic channel state information (CSI). \textcolor{mycolor}{During this phase, we eliminate unnecessary communication links which may not contribute to the performance improvement of collaborative perception in terms of the average transmission latency and assurance of viable communication.} Next, we introduce an adaptive refinement reconstruction approach, which involves creating an adaptive rate-distortion (R-D) strategy that responds to the changing CSI. Here, CAVs send a small subset of raw images to RSUs to enhance data reconstruction and update the encoder and decoder parameters, thus reducing temporal redundancy. Additionally, CAVs convert data into a bit stream using their encoders, which is then sent to the ego CAV. In the final stage, the ego CAV decodes this bit stream and aligns the domain of reconstructed images with its own perceived images' domain. Subsequently, these aligned datasets are combined using a fusion network to create a comprehensive BEV prediction.
\subsection{Transmission Delay Minimization}

In collaborative perception, transmission delay is a critical indicator for CAVs, which is significant for maintaining perception performance and ensuring the driving safety. In order to minimize the transmission delay, we need to model the communication performance in a collaborative perception system. Firstly, we define an adjacent matrix to represent the V2V communication graph, the diagonal elements in this matrix is zero and the off-diagonal elements are set to binary values. The capacity of each communication channel is determined according to the Shannon capacity theorem, taking into account factors like transmit power and channel gain.

A key innovation lies in our adaptive compression approach \cite{fang2024pacp}, which assigns higher priority for data transmission to closer vehicles due to their higher sensing importance. This dynamic adjustment ensures efficient data transmission without losing too much information in the sensing system. The goal is to minimize the average transmission delay across the network by optimizing both the compression ratio and the transmission link matrix, using gradient descent methods for an efficient solution. This method significantly improves data sharing among CAVs, enhancing the overall functionality and safety of autonomous driving systems.


\subsection{Adaptive Refinement Reconstruction}

In this section, we introduce an adaptive refinement reconstruction method aimed at optimizing the R-D trade-off for CAVs. This method dynamically adjusts the compression ratio based on real-time channel conditions, enhancing the efficiency of data transmission in V2V collaborative perception.

Our approach utilizes a convolutional neural network-based encoder and decoder for data compression and reconstruction. The main objective is to minimize the loss function, which comprises two components: the amount of bits needed for compression and the distortion between the original and reconstructed images. The trade-off between these two components is dynamically regulated by an adaptive parameter, which is a function of the compression ratio. This dynamic adjustment allows for flexible adaptation to varying channel conditions, making the R-D trade-off more efficient.

Additionally, we propose a refinement strategy to reduce temporal redundancy in CAV perception data \cite{huAdaptiveCommunicationsCollaborative2023}. This involves a subset of real-time data to train the reconstruction network, enhancing its accuracy. Part of the raw data is sent to a roadside edge server, which then uses it to refine the reconstruction network. This approach leverages historical data from similar scenarios, thereby improving the model's ability to reconstruct future images with greater precision.

\subsection{Domain Alignment}

In collaborative perception for connected and autonomous driving, different CAVs are located in different environments. For example, one CAV may be located in the shade while another CAV is located in the open. In addition, different CAVs are equipped with different cameras that lack unified calibration, which may result in chromatic aberration. To address this issue, we propose a domain alignment method. Specifically, as shown in Fig \ref{fig:overall-architecture}(c), in order to reduce the domain discrepancy between the perceived images of the ego CAV and other CAVs, we first convert the ego vehicle's images to spectrum space by fast Fourier transform (FFT), then decouple it into amplitude spectrum and phase spectrum. We perform the same operations on the other CAVs' images. Then, we align the amplitude spectrum of other CAVs' images with the amplitude spectrum of the ego CAV's images. 
Finally, we convert the aligned spectrum back to spatial space using inverse fast Fourier transform (IFFT). In this way, we can reduce the domain discrepancy between the perceived images of the ego CAV and other CAVs, which can further improve the performance of collaborative perception.

\section{Performance Evaluation} \label{sec:performance-evaluation}

\begin{figure}[t]
  \centering
  \includegraphics[width=0.7\linewidth]{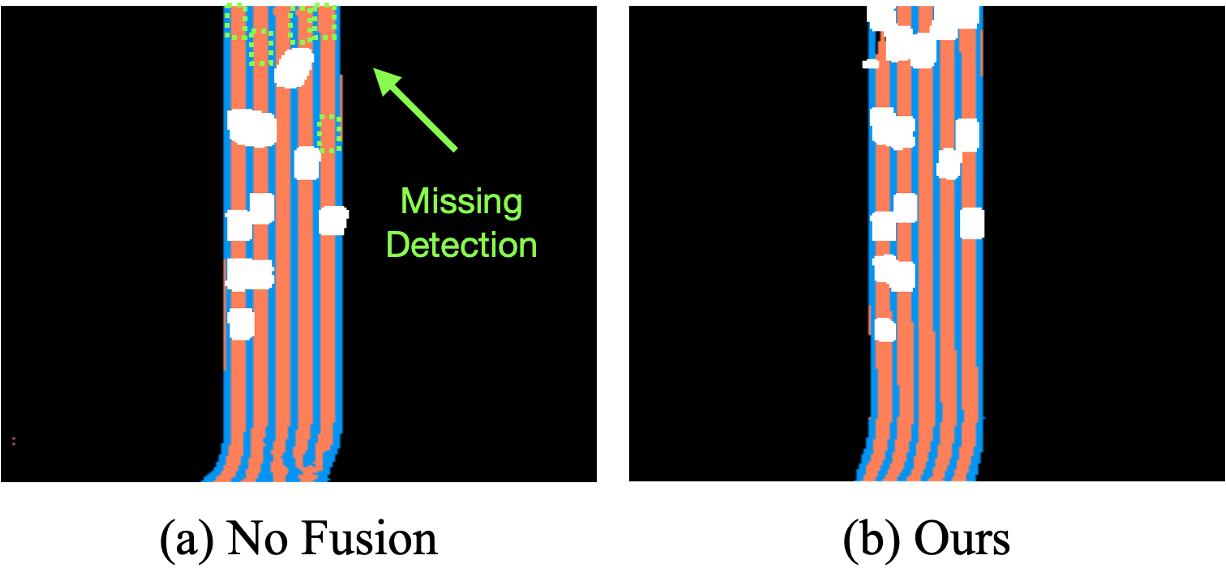}
  \caption{Visualization of BEV segmentation.}
  \label{fig:visualization}
\end{figure}

\begin{table}[t]
    \centering
    \caption{{Performance Comparison} in BEV Segmentation on the OPV2V Camera-Track Dataset.}
    \resizebox{1\columnwidth}{!}{
    \begin{tabular}{p{2cm}|ccccc}
    \hline
    Model        & Road & Lane & Vehicles & Overall  \\ \hline\hline
    No Fusion     & 42.74     & 30.89     &40.73         &   38.12                 \\
    F-Cooper \cite{chenFcooperFeatureBased2019}    & 32.34&      25.01         & 42.27             &33.21                         \\
    AttFuse \cite{xuOPV2VOpenBenchmark2022}    &   43.30&      31.35          & 45.70               &40.11                          \\
    V2VNet \cite{wangV2VNetVehicletoVehicleCommunication2020a}       &     53.00 &         36.11&42.77          & 43.96                         \\
    DiscoNet  \cite{liLearningDistilledCollaboration2021}   &      52.20       &36.19                & 42.97     & 43.48                         \\
    CoBEVT \cite{xuCoBEVTCooperativeBird2023}      &      61.78 & 47.65          & 49.43          & 52.95&                 \\
    Ours &      \textbf{62.60} & \textbf{49.08} & \textbf{53.50} & \textbf{55.06}                          \\ \hline
    \end{tabular}
    }
    \label{tab:baseline comparison}
\end{table}

In order to evaluate our proposed method, we use the OPV2V dataset, a large-scale open dataset for collaborative perception \cite{xuOPV2VOpenBenchmark2022}. Then we build our model by PyTorch and train it on two RTX4090 GPUs utilizing AdamW optimizer. The initial learning rate is $2\times 10^{-4}$ and decays by and exponential factor of $1\times 10^{-2}$. During training and inference, the number of CAVs is set between 2 to 5. \textcolor{mycolor}{In addition, the simulations are based on the 3GPP standard with the following key parameters: local data per vehicle at 40 Mbits, 4 subchannels, computation complexity at 100 cycles/bit, and bandwidth 200 MHz. The vehicle speeds range from 0 to 50 km/h.} In order to evaluate the performance of our proposed method, we utlize the intersection of union (IoU) as the evaluation metric. 

We evaluate our proposed method with several existing methods, which are No Fusion (single vehicle), F-Cooper \cite{chenFcooperFeatureBased2019}, V2VNet \cite{wangV2VNetVehicletoVehicleCommunication2020a}, AttFuse \cite{xuOPV2VOpenBenchmark2022}, DiscoNet \cite{liLearningDistilledCollaboration2021}, and CoBEVT \cite{xuCoBEVTCooperativeBird2023}. These baselines assume that the communication channel is ideal and do not account for domain variations among CAVs. The experimental results, detailed in Table \ref{tab:baseline comparison}, reveal that our method excels in every assessed category. It offers an IoU of 55.06\%, surpassing the nearest competitor by 2.11\%. Notably, in the vehicle class category, which is particularly challenging, our method leads by a margin of 4.07\% over the next best method. These results underscore the superior capability of our method in enhancing the accuracy of collaborative perception within the realm of autonomous driving.

In addition, we also visualize the BEV prediction results of our proposed method with No Fusion, as shown in Fig. \ref{fig:visualization}. We observe that our method can obtain more accurate BEV prediction results compared with No Fusion. No Fusion shows notable deficiencies in accurately identifying vehicles. Our method stands out by effectively and almost flawlessly delineating vehicles, road surfaces, and lanes, including those vehicles that are significantly distant from the ego vehicle. These observations demonstrate the superior performance of our method.

\section{Conclusion}

In this article, we have investigated the research status about V2V collaborative perception, and discussed the challenges about pose errors, asynchronous communications, data volume, etc. 
Then, we have elaborated possible solutions to address these challenges and discussed the future opportunities. In addition, we have leveraged a use case to design an exemplar scheme to deal with the communication efficiency and latency, which is a channel-aware collaborative perception framework to dynamically adjust the communication graph and minimize the average transmission latency. Experiments verify the superiority of our framework compared with the existing state-of-the-art methods.

\bibliographystyle{IEEEtran}
{\small
        \bibliography{ref, ref2}}


\begin{IEEEbiographynophoto}{Senkang Hu} is currently a PhD student with the Department of Computer Science at City University of Hong Kong. His research interests include autonomous driving, vehicle-to-vehicle collaborative perception and smart mobility.
\end{IEEEbiographynophoto}

\begin{IEEEbiographynophoto}
{Zhengru Fang} is currently a PhD student with the Department of Computer Science at City University of Hong Kong. His research interests include collaborative perception, V2X, age of information, and mobile edge computing.
\end{IEEEbiographynophoto}


\begin{IEEEbiographynophoto}
        {Yiqin Deng}  is currently a Postdoctoral Research Fellow with the Department of Computer Science
        at City University of Hong Kong.  Her research interests include edge computing, Internet of Vehicles, and resource management. 
\end{IEEEbiographynophoto}

\begin{IEEEbiographynophoto}{Xianhao Chen} 
is currently 
         an assistant professor with the Department of Electrical and Electronic Engineering, the University of Hong Kong. His research interests include wireless networking, edge intelligence, and machine learning.
\end{IEEEbiographynophoto}

\begin{IEEEbiographynophoto}{Yuguang Fang}
   (S’92, M’97, SM’99, F’08) is currently
  the Chair Professor of Internet of Things with the Department of Computer
  Science at City University of Hong Kong. His research interests include wireless networks, mobile/edge computing, 
  machine learning,
  Internet of Things, and connected and autonomous driving.
\end{IEEEbiographynophoto}


\end{document}